\documentclass{article}

\usepackage{arxiv}

\usepackage{amsmath}
\usepackage{amssymb}
\usepackage{mathtools}
\usepackage{amsthm}

\usepackage[utf8]{inputenc} 
\usepackage[T1]{fontenc}    
\usepackage{hyperref}       
\usepackage{url}            
\usepackage{booktabs}       
\usepackage{amsfonts}       
\usepackage{nicefrac}       
\usepackage{microtype}      
\usepackage[capitalize,noabbrev]{cleveref}       
\usepackage{lipsum}         
\usepackage{graphicx}
\usepackage{doi}

\usepackage{natbib}
\renewcommand{\cite}[1]{\citep{#1}}
\setcitestyle{authoryear,round,citesep={;},aysep={,},yysep={;}}

\title{Learning Generalized Zero-Shot Learners for Open-Domain Image Geolocalization}

\date{January 30, 2023}

\author{Lukas Haas\\
	Department of Computer Science\\
	Stanford University\\
	\texttt{lukashaas@cs.stanford.edu} \\
	\And
	Silas Alberti\\
	Department of Electrical Engineering\\
	Stanford University\\
	\texttt{salberti@stanford.edu} \\
	\And
        Michal Skreta\\
	Department of Computer Science\\
	Stanford University\\
	\texttt{michal.skreta@stanford.edu} \\
}

\renewcommand{\headeright}{}
\renewcommand{\undertitle}{Preprint}
\renewcommand{\shorttitle}{Learning Generalized Zero-Shot Learners for Open-Domain Image Geolocalization}

\hypersetup{
pdftitle={StreetCLIP Preprint},
pdfsubject={q-bio.NC, q-bio.QM},
pdfauthor={Lukas Haas, Silas Alberti, Michal Skreta},
pdfkeywords={Contrastive Pretraining, Generalized Zero-Shot, Zero-Shot Learning, Image Geolocalization, Visual Geolocation Recognition, CLIP, Computer Vision, Multi-Modal},
}

\begin{document}
\maketitle

\begin{abstract}
Image geolocalization is the challenging task of predicting the geographic coordinates of origin for a given photo. It is an unsolved problem relying on the ability to combine visual clues with general knowledge about the world to make accurate predictions across geographies. We present \href{https://huggingface.co/geolocal/StreetCLIP}{StreetCLIP}, a robust, publicly available foundation model not only achieving state-of-the-art performance on multiple open-domain image geolocalization benchmarks but also doing so in a zero-shot setting, outperforming supervised models trained on more than 4 million images. Our method introduces a meta-learning approach for generalized zero-shot learning by pretraining CLIP from synthetic captions, grounding CLIP in a domain of choice. We show that our method effectively transfers CLIP's generalized zero-shot capabilities to the domain of image geolocalization, improving in-domain generalized zero-shot performance without finetuning StreetCLIP on a fixed set of classes.
\end{abstract}

\vspace{1em}


\section{Introduction}

Image geolocalization touches many aspects of our lives with applications in search engines and on-device photo tagging serving billions of users every day. By understanding the hidden locational clues in images, entirely new approaches of analyzing the natural and built environment are being opened up with profound implications for a number of fields, ranging from the recognition of weather, season, and climate patterns to rural and urban scene understanding, and improvements in navigation and self-driving car technology. Since the beginning of 2022, image geolocalization has additionally garnered extensive media coverage for becoming an immediate priority of investigative journalists and open source intelligence (OSINT) researchers in their attempt to verify information and to document war atrocities in Ukraine, extracting geolocational information from social media content.

Despite high academic and public interest, image geolocalization remains an extremely challenging problem. This is because training datasets are geographically sparse, often limited to specific countries, and biased towards urban or rural scenes. The task is further complicated by the fact that geolocalization requires reasoning on multiple levels of geographic granularity (e.g. countries, cities, and neighborhoods) and that the geolocation of an image is a property that can often not be observed directly. Effective image geolocalization thus forces a model to not only distill critical information from subtle visual clues but to also combine these clues with a general understanding of the world, including abstract concepts such as weather patterns, political boundaries, or on what side of the road people are driving on.

\begin{figure*}[htb]
\vskip 0.1in
\begin{center}
\centerline{\includegraphics[height=6cm]{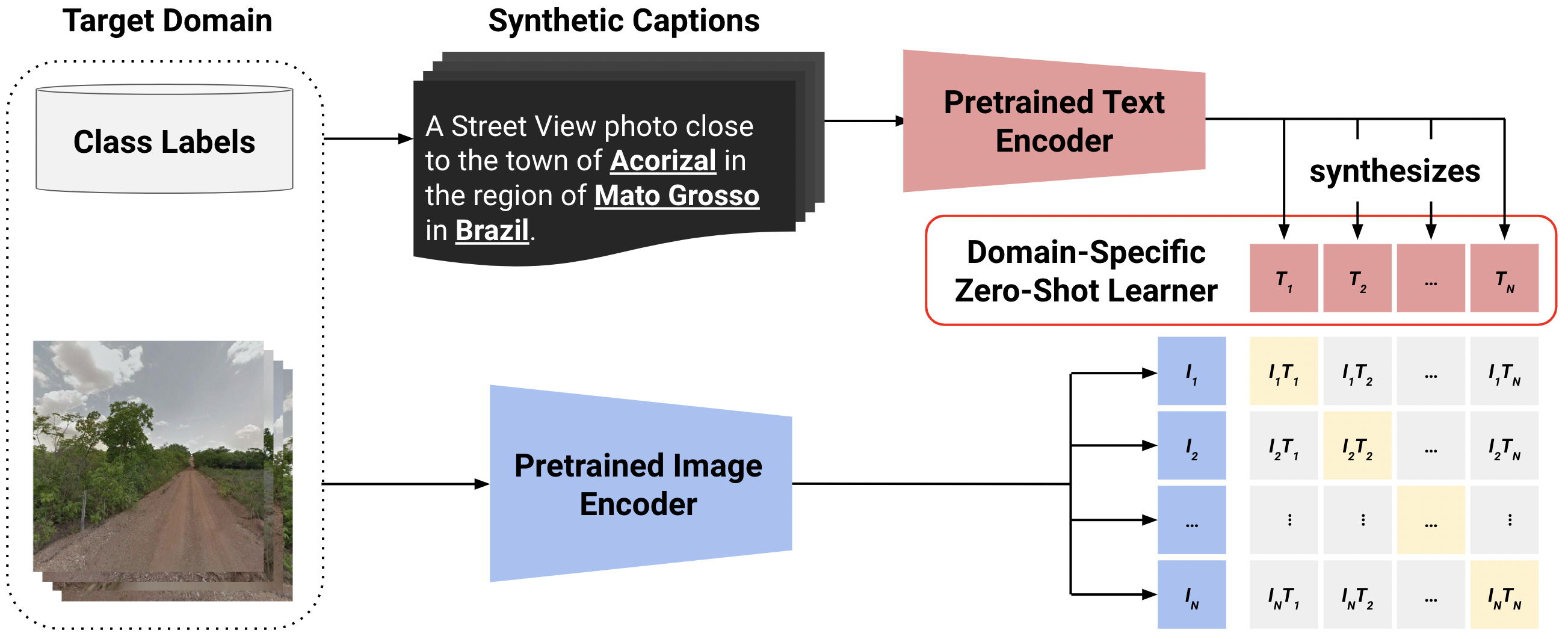}}
\caption{\textbf{StreetCLIP's Synthetic Caption Pretraining}. We formulate the task of image geolocalization in natural language via synthetic captions at
various levels of geographic granularity. For every batch, our model synthesizes a generalized zero-shot learner, thus learning how to zero-shot learn within a specific domain. The figure layout draws on \citet{radford21a}.}
\label{streetclip}
\end{center}
\vskip -0.1in
\end{figure*}

Prior works on image geolocalization have employed convolutional neural networks \cite{Weyand_2016, vo_et_al} and more recently transformer models \cite{pramanick_2022, wu_huang_2022, luo_2022}, but have all struggled to generalize across geographies. In \citet{wu_huang_2022}, the authors note that CLIP \cite{radford21a} has impressive zero-shot capabilities which extend to image geolocalization and that finetuning a model based on CLIP's image embeddings can further increase performance. However, when evaluated on data that is out of distribution, the performance degrades to be worse than the original CLIP model's accuracy \cite{wu_huang_2022}. This raises the question of how we can equip models with more robust knowledge, enabling both transfer and zero-shot learning to unseen geographies and similarly structured tasks.

Motivated by this research question, we introduce a domain-specific pretraining method that generates synthetic captions for contrastive learning, enabling the use of natural language to ground CLIP in the context of the image geolocalization task. For a given image classification task, we uses class labels to derive synthetic captions from a domain-specific caption template. Together with the corresponding images, the synthetic captions are then used for an additional pretraining round of CLIP in a contrastive setting. We show that our method is equivalent to generating a new domain-specific generalized zero-shot learner during every batch iteration which learns to distinguish between classes seen and unseen during training. In doing so, we show that our method is a meta-learning approach for generalized zero-shot learning, encouraging models to learn how to synthesize better domain-specific generalized zero-shot learners.

To demonstrate the effectiveness of domain-specific pretraining method via our synthetic caption method, we introduce StreetCLIP, a robust image geolocalization model trained on an original dataset of 1.1 million Street View images. StreetCLIP achieves state-of-the-art (SOTA) performance on the open-domain image geolocalization benchmarks IM2GPS \cite{hays_2008} and IM2GPS3K \cite{vo_et_al}, improving the geolocation prediction accuracy for some distance thresholds between 0.3 and 2.4 percentage points. Beyond improving upon SOTA performance, our results are notable because in contrast to prior SOTA methods, StreetCLIP performs inference using zero-shot learning, outperforming supervised models trained on more than 4 million in-distribution images. We make StreetCLIP available to the broader research community by releasing our pretrained model on Hugging Face\footnote{StreetCLIP is publicly available under the CC-BY-NC-4.0 license at \href{https://huggingface.co/geolocal/StreetCLIP}{https://huggingface.co/geolocal/StreetCLIP}.}.

Finally, because our domain-specific pretraining method does not rely on any image captioning datasets, it can be extended to any image classification problem conditional on class names being expressible in natural language. This opens the door for further investigation whether our method can also improve CLIP's generalized zero-shot learning capabilities in other domains. 

\section{Related Work}

\subsection{Image Geolocalization}

The task of image geolocalization, also known as visual place recognition (VPR), is a difficult problem due to the sheer diversity of geographies and conditions in which images are taken. Because datasets suitable for geolocational analysis are often heavily biased towards certain geographies, prior research has primarily focused on image geolocalization in constrained environments.

Despite the increasing number of publications in the field, the research community has so far failed to clearly distinguish between problem formulations that aim to geolocate images based on a limited, fixed set of classes or geographies (e.g. landmarks or specific cities), and a setting in which a model must reasonably expect that a test set image could have been taken anywhere in the world. For instance, \citet{Berton_2022_CVPR} introduces a benchmark for image geolocalization based on six different datasets, yet five of them evaluate image geolocalization within a single city or suburb, while the last dataset, Mapillary \cite{warburg_2022}, draws on images from a fixed set of 30 cities. While the benchmark introduced by \citet{Berton_2022_CVPR} is an important contribution to the field of image geolocalization, it is limited in its practicality for real-world applications and does not evaluate the transfer-learning or zero-shot capabilities required by planet-scale image geolocalization without strong priors about the distribution of test set images. 

To draw a clear distinction between these two problem formulations, we introduce the terms closed-domain and open-domain image geolocalization. The objective of this distinction is to improve the evaluation process of image geolocalization models and to allow for a better specification of their intended use cases.

\subsubsection{Closed-Domain Image Geolocalization}

Closed-domain image geolocalization (CDIG) is the problem of predicting the location of images either from a fixed set of geolocation classes or within a geographic region such as a selection of cities or countries. Because of the limited availability of comprehensive datasets and computational resources, older related work developed specialized, feature-based approaches constrained to either specific natural environments such as mountain ranges \cite{baatz_2012, saurer_2016}, deserts \cite{tzeng_2013}, or beaches \cite{cao_2012}, or the built environment of single cities \cite{zamir_shah_2010, zamir_shah_2014}. With the advent of deep learning in computer vision, CDIG performance has improved significantly both on street-level \cite{berton_2022a} and broader geographic scales \cite{suresh_2018}. The concurrent work of \citet{wu_huang_2022} and \citet{luo_2022} builds on top of these approaches, being the first to apply CLIP \cite{radford21a} to the problem of CDIG. \citet{wu_huang_2022} additionally are the first to employ CLIP in a zero-shot setting via linear probing. In contrast to \citet{wu_huang_2022}, our work uses a planet-scale, hierarchical linear probing strategy that enables zero-shot image classification models to perform open-domain image geolocalization.

\subsubsection{Open-Domain Image Geolocalization}

Open-domain image geolocalization (ODIG) does not restrict the geographic domain of test set images, meaning models have to perform image geolocalization in an unconstrained manner. The first modern attempt at planet-scale image geolocalization is attributed to IM2GPS \cite{hays_2008} with follow-up work by \citet{zamir_shah_2014} and \citet{vo_et_al}. All three approaches rely on image retrieval methods from large reference datasets during test time which is slow and computationally intensive.

In 2016, Google researchers released PlaNet \cite{Weyand_2016} that first applied convolutional neural networks \cite{krizhevsky_2012} in an end-to-end fashion to photo geolocalization. It also first cast the problem as a classification task after \citet{de_brebisson_2015} had demonstrated that it was difficult for models to directly predict geographic coordinates. Further work included the incorporation of scene information \cite{budack_2018} and the progression to vision transformer architectures \cite{pramanick_2022} based on the work of \citet{vaswani_2017}. While all these approaches achieved impressive results, their supervised classification setups transform ODIG into CDIG problems during training, limiting performance under distribution shifts.

Our method solves both the limitations of retrieval-based and supervised classification methods by being the first work to apply zero-shot learning to the problem of open-domain image geolocalization.

\subsection{Learning Under Distribution Shifts} 

Successful open-domain image geolocalization requires models to be robust to distribution shifts and ideally to also perform well on out-of-distribution data, for example on countries not seen during training. ODIG is thus a perfect environment to evaluate both the robustness and zero-shot capabilities of models with learnings from ODIG extending to other domains. CLIP, introduced by \citet{radford21a}, has been shown to be a robust image classification model with exceptional zero-shot capabilities by employing natural language supervision. A key question in the literature has consequently emerged: how can we transfer CLIP's knowledge to a specific target domain?

In their work on image geolocalization, \citet{wu_huang_2022} note that finetuning a model based on CLIP's image embeddings is an effective transfer learning approach conditional on test set data being drawn from the same distribution as used during finetuning. The authors further observe that finetuning hurts performance compared to zero-shot learning with CLIP when evaluating on a test set drawn from a distribution different than their finetuning distribution.

The phenomenon of finetuning deteriorating model robustness to distribution shifts has also been observed in a broader image classification context \cite{Wortsman_2022_CVPR}. \citet{Wortsman_2022_CVPR} address this problem by ensembles the weights of the original zero-shot and finetuned models, achieving large accuracy gains under distribution shifts. In contrast to \citet{Wortsman_2022_CVPR}, our approach is capable of not only improving performance under distribution shifts, but also on out-of-distribution data which finetuning methods cannot achieve because of a fixed set of classes.

Related work on near out-of-distribution learning demonstrates that the pretraining procedure of large transformers is responsible for their robustness to distribution shifts. This is because pretraining on different datasets reduces the models' vulnerability to shortcut learning \cite{geirhos_2020, fort_2021}. Leveraging this insight, \citet{shen_2021} use masked language modeling as an additional pretraining step for CLIP, achieving better generalization and downstream task performance.

Despite the success of contrastive pretraining in creating CLIP, transferring CLIP's knowledge to a target domain via supervised contrastive learning \cite{khosla_2020} remains largely unexplored, likely due to the lack of domain-specific image-caption datasets \cite{clip_rsicd}. Our method addresses this limitation via synthetic captions derived from training class labels. While supervised contrastive learning has been shown to produce degenerate representations, mapping all instances of a class to the same point in latent space \cite{chen_2022d}, this is not the case for multi-modal models supervised via natural language which have a virtually infinite number of possible classes. Because semantically similar captions can be represented with many different tokens, natural language supervision in CLIP can be understood as a form of label smoothing, preventing the class collapse observed in \citet{chen_2022d}.

\subsection{Generalized Zero-Shot Learning}

Taking robustness to domain shifts to the extreme, an ideal property of ODIG models is the ability to correctly classify images from countries, regions, or cities not seen during training, known as zero-shot learning in learning theory. In its simplest form, zero-shot learning is the task of learning a classifier $f : X \rightarrow Y$ such that $f$ can correctly predict novel values of $Y$ not seen during training \cite{palatucci_2009}. This framework can be extended to generalized zero-shot learning (GZSL) in which $Y$ includes both seen and unseen classes during inference.

While the literature on GZSL for multi-modal models is fairly novel, it identifies two important gaps; first, generalized zero-shot learning in the context of CLIP requires further investigation \cite{pourpanah_2022}, and second, most GZSL methods are based on ideal datasets with learnings not translating to real-world datasets \cite{pourpanah_2022}. Our work addresses both of these research gaps by introducing a novel pretraining method for CLIP to improve GZSL while evaluating it on two challenging real-world benchmarks: IM2GPS \cite{hays_2008} and IM2GPS3K \cite{vo_et_al}.

\section{Preliminaries}
\label{sec:prelimiaries}

In this section, we lay out the notation and existing theory behind generalized zero-shot learning and how CLIP's zero-shot capabilities relate to it. This will be necessary to show how our method trains CLIP to learn how to perform generalized zero-shot learning.

\subsection{Generalized Zero-Shot Learning}

Generalized zero-shot learning is the task of training a classifier to correctly predict both classes which were seen and unseen during training. Formally, let $Y^{s} = \{y^s_1, \ldots, y^{s}_{N_s}\}$ be the set of all classes observed during training and $Y^{u} = \{y^u_1, \ldots, y^{u}_{N_u}\}$ the set of all unseen classes, with $Y^s \cap Y^u = \emptyset$, $N_s$ being the number of seen classes and $N_u$ the number of unseen classes. Further, let $Y = Y^s \cup Y^u$ and $X$ be the set of all possible model inputs and $X^{tr} \subset X, X^{ts} \subset X$ be the set of training and testing inputs, respectively, with $X^{tr} \cap X^{ts} = \emptyset$.

For a classification setting, meaning there exist functions $g^{tr}$ and $g^{ts}$ which map training and testing examples to their single ground truth class, respectively, we can now define the training and testing datasets for GZSL as follows: $$\mathcal{D}^{tr} = \{(x_i, y_i) | x_i \in X^{tr}, y_i \in Y^s, g^{tr}(x_i) = y_i\},$$ and $$\mathcal{D}^{ts} = \{(x_i, y_i) | x_i \in X^{ts}, y_i \in Y, g^{ts}(x_i) = y_i\}.$$

In contrast to zero-shot learning which would attempt to learn a classifier $f_{\text{ZS}} : X \rightarrow Y^u$, the objective of GZSL is to learn a classifier $f_{\text{GZS}} : X \rightarrow Y$ which correctly classifies examples from $D^{ts}$ having only access to $D^{tr}$ during training. Because $Y^u \subset Y$, GZSL can be seen as a generalization of traditional zero-shot learning.

\subsection{How CLIP Performs Zero-Shot Learning}
\label{sec_clip_math}

CLIP can be abstracted to consist of four distinct functions: a text encoder $f : X \rightarrow \mathcal{X}$, mapping a batch of captions in natural language $x \in X$ to a batch of representations in the latent space $\mathcal{X}$, an image encoder $g : V \rightarrow \mathcal{V}$, similarly mapping a batch of images $v \in V$ to a batch of representation in the latent space $\mathcal{V}$, and two prediction functions $h, j : \mathcal{X} \times \mathcal{V} \rightarrow \mathbb{R}^{N^{\text{batch}} \times N^{\text{batch}}}$ mapping $N^{\text{batch}}$ latent text and image representations to a matrix of probabilities signifying which caption and image representations correspond to each other. The functions $h$ and $j$ simply compute the matrix product of their inputs and a softmax, where $h$ computes the softmax over the dimension of text representations and $j$ over the image representations.

\citet{radford21a} notes that during inference, CLIP can be understood to synthesize a zero-shot learner to classify a given image. To show this mathematically, let $v \in V$ be an image batch consisting only of a single image, and $x \in X$ be $N^{\text{ZS}}$ natural language captions from a specific template such as \texttt{"An image of an \{object\}"}. Passing these inputs through CLIP's image and text encoders yields $\textbf{v} = g(v)$ and $\textbf{X} = f(x)$ where $\textbf{v} \in \mathbb{R}^{d \times 1}$ and $\textbf{X} \in \mathbb{R}^{N^{\text{NZ}} \times d}$, with $d$ being the dimension of CLIP's joint embedding space. Given that $x$ was derived from a prompting template, each row of $\textbf{X}$ encodes the semantics of a specific \texttt{\{object\}}. 

Further, let $h$ be the prediction function from CLIP's training procedure which computes the softmax over the dimension of text embeddings. If $\textbf{v}$ is a single image representation, then $h$ is a linear classifier parameterized by $\textbf{X}$: 

\begin{align}
    h(\textbf{v}; \textbf{X}) = \frac{\textbf{X}\textbf{v}}{\sum_{j=1}^{N^{\text{ZS}}} \textbf{X}_{j}^{\top} \textbf{v}},\label{eq_clip_0}
\end{align}

computing a softmax over the matrix-vector product $\textbf{X}\textbf{v}$. The output of $h(\textbf{v})$ is a probability vector over caption representations, each entry corresponding to a specific \texttt{\{object\}}.

We note that the labels, or \texttt{\{object\}}s, supplied via the caption template and used to generate the weight matrix $\textbf{X}$ can be chosen arbitrarily whether seen or unseen during training. Since $\textbf{X}$ fully parameterizes a linear classifier, instantiating $h(\cdot)$ with seen \emph{and} unseen labels is equivalent to CLIP synthesizing a generalized zero-shot learner during inference. 

\section{Learning How to Zero-Shot Learn for Open-Domain Image Geolocalization}

The mathematical intuition behind CLIP's zero-shot capabilities and gaps in the literature exposes two intriguing questions: how can CLIP learn to synthesize better generalized zero-shot learners, and how can we learn to create domain-specific learners, capable of classifying new classes of images for which no training data was available? We address these two questions using image geolocalization as a comprehensive case study, emphasizing that our method is general and could be applied to other domains.

\subsection{Synthetic Caption Domain-Specific Pretraining}
\label{sec:scdp}

We introduce synthetic caption domain-specific pretraining as an additional pretraining step for CLIP both to learn how to train better zero-shot learners as well as to develop domain-specific zero-shot capabilities. In section \ref{sec_clip_math}, we described how CLIP synthesizes generalized zero-shot learners during inference – synthetic caption domain-specific pretraining extends this paradigm to the training phase, explicitly training CLIP to learn better generalized zero-shot learners.

Building on top of a pretrained CLIP model, our method performs an additional domain-specific pretraining round, depicted in \cref{streetclip}. For every training set image, we generate a synthetic image caption by formulating the task of image geolocalization in natural language using the following prompting template: 

\begin{verbatim}
A Street View photo close to the town of {city} in the region of {region} in {country}.
\end{verbatim}
\label{tmp:training}

The tuple (\texttt{\{city\}}, \texttt{\{region\}}, \texttt{\{country\}}) denotes the location where a specific image was taken and is the class label our model learns to predict. We include multiple levels of geographic granularity in the template for two reasons. First, city names can be ambiguous, and second, by employing our geographic hierarchy, a model can learn similar representations for cities within the same country or region.

Because we use the same prompting template for all samples during training, our synthetic captions only differ in the semantics of the class labels. This means that for every training batch of size $N^{\text{batch}}$, CLIP's text encoder synthesizes a weight matrix $\textbf{X} \in \mathbb{R}^{N^{\text{batch}} \times d}$ of which every row encodes a class label of a specific sample in the batch, $d$ being the dimension of CLIP's embeddings space. $\textbf{X}$ is thus the weight matrix of a new, domain-specific generalized zero-shot learner for every batch iteration.

To demonstrate how our synthetic caption domain-specific pretraining method optimizes for creating better generalized zero-shot learners, we refer to CLIP's loss function. \citet{radford21a} derive two individual cross-entropy loss terms by computing the matrix product between the text and vision encoder embeddings and then taking a softmax over the dimension of text and image embeddings, resulting in $\mathcal{L}_{\text{Text}}$ and $\mathcal{L}_{\text{Images}}$, respectively. CLIP's loss consequently becomes:
\begin{align}
    \mathcal{L}_{\text{CLIP}} = 0.5 \cdot (\mathcal{L}_{\text{Text}} + \mathcal{L}_{\text{Images}})
\end{align}

However, if we use class labels to fill a pre-defined caption template during training, new intriguing properties of the loss terms emerge. By the mathematical intuition laid out in section \ref{sec:prelimiaries}, $\mathcal{L}_{\text{Text}}$ now directly optimizes for reducing the cross-entropy loss of a linear generalized zero-shot learner which takes the vision encoder features as input. Likewise, the second loss term $\mathcal{L}_{\text{Images}}$ employs a cross-entropy loss to now optimize image representations to better correspond to a specific class label, attempting to adjust the vision encoder's weights to render zero-shot learning easier. As a result, CLIP's loss function can be reformulated as:

\begin{align}
    \mathcal{L}_{\text{CLIP}} = 0.5 \cdot (\mathcal{L}_{\text{GZSL}} + \mathcal{L}_{\text{Vision Representation}})
    \label{eq:loss_clip_gzsl}
\end{align}

where $\mathcal{L}_{\text{GZSL}}$ is the cross-entropy loss of the batch's linear generalized zero-shot learner, and $\mathcal{L}_{\text{Vision Representation}}$ is the cross-entropy loss of optimizing image embeddings to correspond to a specific class label.

The main contribution of our synthetic caption domain-specific pretraining method is that equation \ref{eq:loss_clip_gzsl} corresponds to a loss that learns to generate better generalized zero-shot learners – a meta-learning procedure for generalized zero-shot learning. The fact that this procedure is domain-specific allows CLIP's zero-shot capabilities to be improved within a specific domain.

\subsection{A Planet-Scale Image Geolocalization Dataset}
\label{sec:datasetours}

In evaluating the suitability of datasets for our work, we aimed for our training dataset to fulfill three specific criteria. First, our dataset should be planet-scale – a requirement for sufficiently high ODIG performance. Second, our dataset should come from a different geographical distribution than common image geolocalization benchmarks to demonstrate our model's transfer and zero-shot learning capabilities. Finally, our dataset should not overlap with either YFCC100M \cite{thomee_2016}, on which CLIP \cite{radford21a} was trained on, or with any image geolocalization benchmark datasets.

These three criteria led us to the decision to collect an original dataset of Street View images to demonstrate the effectiveness of synthetic caption domain-specific pretraining within the realm of open-domain image geolocalization. To that end, we obtain 275,000 coordinate pairs for which Google Street View images are available from Geoguessr, a Swedish company developing a popular geolocation guessing game. For each coordinate pair, we collect four images together covering a full 360-degree view, resulting in an original dataset of 1.1 million Google Street View images. To obtain the \texttt{\{city\}}, \texttt{\{region\}}, and \texttt{\{country\}} labels for each image, we employ the open-source reverse-geocoding service Nominatim.

Our final planet-scale image geolocalization dataset includes images from 101 countries, the United States making up the largest share of images with a share of 1.92\%. More details can be found in Appendix \ref{app:trainingdataset}.

\subsection{Training}

Our work employs the synthetic caption domain-specific pretraining method laid out in section \ref{sec:scdp} to train our model StreetCLIP\footnote{Available online: \href{https://huggingface.co/geolocal/StreetCLIP}{https://huggingface.co/geolocal/StreetCLIP}.}, a robust, publicly available foundation model for open-domain image geolocalization. Before we perform our domain-specific pretraining, we initialize StreetCLIP with the weights of OpenAI's own pretrained version of CLIP \cite{clip_large} using 14x14 pixel patches to transform images with a 336-pixel side length into a sequence of 576 image patches as input to its vision encoder transformer.

During training, we use a batch size of 32 and an AdamW optimizer \cite{loshchilov_2017} with a learning rate of $1e^{-6}$. We generate synthetic image captions from the template described in section \ref{tmp:training} and the reverse-geocoded class names for the image's \texttt{\{city\}}, \texttt{\{region\}}, and \texttt{\{country\}}. All input images are preprocessed in the same way as for OpenAI's pretrained CLIP version. A more detailed description of the training parameters can be found in Appendix \ref{app:training}.

\subsection{Hierarchical Linear Probing}
\label{sec:probing}

\begin{figure}[htb]
\vskip 0.2in
\begin{center}
\centerline{\includegraphics[height=6.5cm]{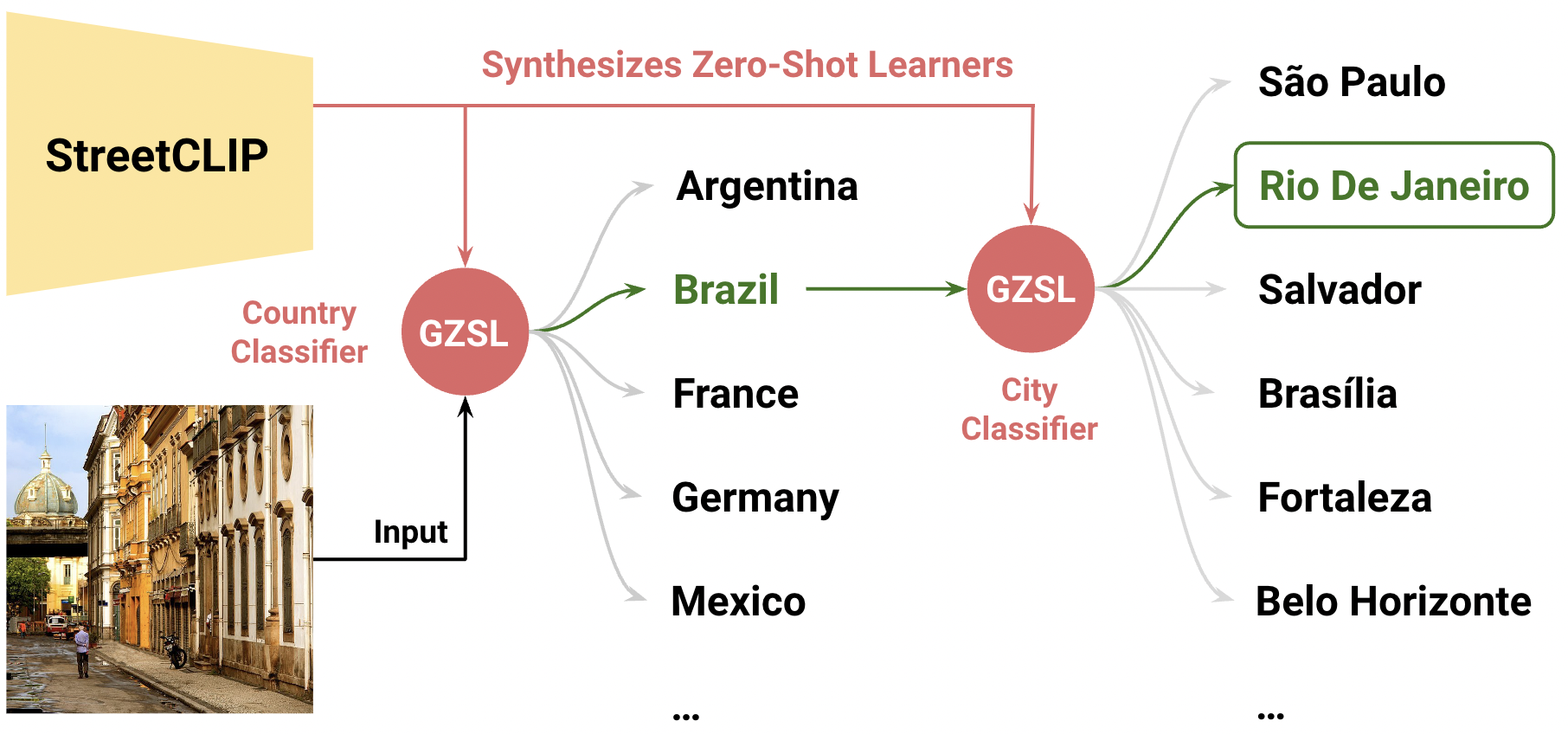}}
\caption{\textbf{Hierarchical Linear Probing Strategy}. During inference, StreetCLIP synthesizes both a country-level and a city-level generalized zero-shot learner using two different caption templates. Given an input image, our method first identifies the country it deems to be the most likely image origin and then refines its guess within that country's 30 most populous cities.}
\label{fig:linear_probe}
\end{center}
\vskip -0.1in
\end{figure}

To demonstrate StreetCLIP's zero-shot image geolocalization capabilities, we devise a hierarchical linear probing strategy: we probe for the correct country first before making predictions at a more granular level. Because of the vast number of cities in the world, a hierarchical strategy significantly speeds up the inference process while also resulting in better performance due to eliminating the risk of city name ambiguity, for example predicting London, Canada instead of London, UK. 

Figure \ref{fig:linear_probe} describes the hierarchical linear probing procedure in detail. During inference, a given image is passed through two generalized zero-shot learners synthesized by StreetCLIP to sequentially determine the most likely country and city of the image's origin. This is achieved via a caption template for countries \texttt{"A Street View photo in \{country\}."} and one for cities \texttt{"A Street View photo from \{city\}."}.

We use a comprehensive list of the world's countries\footnote{See Appendix \ref{app:listcountries}.} and a list of cities derived from SimpleMaps\footnote{See Appendix \ref{app:listcities}.} to generate a decision space for our linear probes. Because disproportionately many images in our training dataset stem from the United States, for our country linear probe, we replace the United States class label with corresponding state-level labels using the template \texttt{"A Street View photo in \{state\}, United States."}.

\section{Experiments}

The central hypothesis of our work is that synthetic caption domain-specific pretraining improves CLIP's generalized zero-shot capabilities applied to a specific task. We evaluate our method in the context of open-domain image geolocalization because it is an extremely challenging task, forcing a model to adapt to strong domain shifts and reason about the world by skillfully combining visual clues and abstract concepts. 

\begin{table*}[t]
\caption{Evaluation of StreetCLIP on Open-Domain Image Geolocalization Benchmarks}
\label{table:ODIG}
\vskip 0.15in
\begin{center}
\begin{small}
\begin{sc}
\begin{tabular}{llcccc}
\toprule
& & \multicolumn{4}{c}{\textbf{Distance (\% @ km)}}\\
\textbf{Benchmark} & \textbf{Model} & \textit{City} & \textit{Region} & \textit{Country} & \textit{Continent} \\
& & 25km & 200km & 750km & 2,500km\\
\midrule
  \textbf{IM2GPS}  & PlaNet \cite{Weyand_2016} & 24.5 & 37.6 & 53.6 & 71.3\\
 $n=237$ & ISNs \cite{budack_2018} & 43.0 & 51.9 & 66.7 & 80.2\\
  &      TransLocator \cite{pramanick_2022}  & \textbf{48.1} & \textbf{64.6} & \textbf{75.6} & 86.7       \\
 \cmidrule{2-6}
                    &      Zero-Shot CLIP (ours)       &  27.0 & 42.2 & 71.7 & 86.9      \\
                    &      Zero-Shot StreetCLIP (ours)    &  28.3  & 45.1 & 74.7 & \textbf{88.2}       \\
\cmidrule{2-6}
                    & \textbf{$\Delta_{\text{StreetCLIP - CLIP}}$} & +1.3 & +2.9 & +3.0 & +1.3\\
\midrule
\textbf{IM2GPS3K} & PlaNet \cite{Weyand_2016} & 24.8 & 34.3 & 48.4 & 64.6\\
$n=2997$ & ISNs \cite{budack_2018} &  28.0 & 36.6 & 49.7 & 66.0\\
& TransLocator \cite{pramanick_2022} & \textbf{31.1} &  \textbf{46.7} & 58.9 &  80.1       \\
 \cmidrule{2-6}
                    &      Zero-Shot CLIP (ours)      &  19.5 & 34.0 & 60.0 & 78.1       \\
                    &      Zero-Shot StreetCLIP (ours)    &  22.4  & 37.4 & \textbf{61.3} & \textbf{80.4}       \\
\cmidrule{2-6}
                    & \textbf{$\Delta_{\text{StreetCLIP - CLIP}}$} & +2.9 & +3.4 & +1.3 & +2.3\\
\bottomrule

\end{tabular}
\end{sc}
\end{small}
\end{center}
\vskip -0.1in
\end{table*}

\subsection{Benchmark Datasets}

Despite a variety of evaluation datasets existing for the problem of image geolocalization, most datasets relate to closed-domain image geolocalization, testing model performance in constrained environments with limited real-world applicability \cite{Berton_2022_CVPR}. 

The only two planet-scale, widely used open-domain image geolocalization datasets are IM2GPS \cite{hays_2008}, containing 237 test set images from 78 countries and IM2GPS3K \cite{vo_et_al} with 2997 images from 112 countries. While IM2GPS's geo-tagged images were collected from all across the internet with famous landmarks being over-represented, IM2GPS3K is a collection of images from Flickr, similar but not overlapping with the YFCC100M \cite{thomee_2016} dataset CLIP was trained on. The distribution of both datasets is very different from StreetCLIP's training dataset described in section \ref{sec:datasetours} both geographically and in content. This enables us to test StreetCLIP's performance out-of-distribution.

While other ODIG datasets exist derived from YFCC100M \cite{thomee_2016}, these datasets are overlapping with OpenAI's CLIP implementation's training data and are thus not applicable to our evaluation process.

\subsection{Experimental Settings}
We perform the evaluation of both CLIP \cite{clip_large} and StreetCLIP on the benchmark datasets IM2GPS and IM2GPS3K using our hierarchical linear probing strategy described in section \ref{sec:probing}. The evaluation of our work is done entirely using zero-shot learning, making for an interesting comparison to work in prior literature which includes models trained on millions of images from a similar distribution to our benchmark datasets.

The objective for our benchmark datasets is to predict the images' coordinates of origin with as little deviation as possible. To ensure comparability, we evaluate our work following the set of metrics set forth in prior literature: given a distance in kilometers between the predicted coordinates to the ground truth coordinates, what percentage of test set coordinate distances are below a certain kilometer threshold? This metric is called Percentage at Kilometer (\% @ KM). We follow the conventions of thresholds from the literature, using 25, 200, 750, and 2,500 kilometer thresholds while leaving out the 1 kilometer threshold as our method does not make predictions at a more granular level than cities.

To generate a prediction, we perform hierarchical linear probing to guess a city within a country and use the SimpleMaps dataset (Appendix \ref{app:listcities}) to transform the city name into geographic coordinates. Finally, we use the Haversine formula to get an accurate estimate of the distance between our prediction and the ground truth coordinates in kilometers.

\subsection{Results}
Table \ref{table:ODIG} shows the performance of CLIP and StreetCLIP on the two selected open-domain image geolocalization benchmarks using hierarchical linear probing.

Our model StreetCLIP achieves state-of-the-art (SOTA) performance on both benchmark datasets on a total of three distance thresholds, notably using zero-shot learning. On IM2GPS, StreetCLIP outperforms the prior SOTA performance of TransLocator by \citet{pramanick_2022} for the 2,500 kilometer threshold by 1.5 percentage points, with worse performance than TransLocator on lower kilometer thresholds. On the larger benchmark dataset IM2GPS3K, StreetCLIP sets a new SOTA performance on two distance thresholds, beating the SOTA on the 750 kilometer threshold by 2.4 percentage points and on the 2,500 kilometer threshold by 0.3 percentage points. Again, StreetCLIP performs worse than TransLocator on lower kilometer thresholds. 

Further, we observe that domain-specific pretraining with our synthetic caption method drives a substantial improvement in image geolocalization performance compared to our zero-shot CLIP model (an ablation of our pretraining method): for all distance thresholds, our synthetic caption pretraining method improves performance between 1.3 and 3.4 percentage points. This demonstrates that our method improved CLIP's zero-shot reasoning capabilities within the domain of image geolocalization, despite training on a dataset of Street View images which are considerably different from user-uploaded images on Flickr. 

\subsection{Discussion and Limitations}

The results of our experiments demonstrate that our synthetic caption pretraining method is capable of significantly improving CLIP's generalized zero-shot capabilities applied to a specific task while achieving SOTA performance on a selection of benchmark metrics. These results, however, must be placed within the context of our zero-shot learning setup. Notably, StreetCLIP's performance is achieved via planet-scale linear probing in zero-shot in contrast to TransLocator which was trained in a supervised fashion on more than 4 million images. Furthermore, while TransLocator was trained on geo-tagged images from Flickr just as our benchmark dataset IM2GPS3k, StreetCLIP realizes its performance being 
pretrained on a dataset of StreetView images that experience a strong domain shift to our benchmarks.

Observing the results in Table \ref{table:ODIG}, it becomes clear that hierarchical probing with both CLIP and StreetCLIP works especially well on higher kilometer thresholds. A limitation of our approach, however, is the accurate prediction of image geolocations on more granular kilometer thresholds such as the city or region level. StreetCLIP and CLIP significantly underperform on more granular prediction levels, likely because landmarks are not included in our probing procedure although common in the benchmarks, because our models can only predict the center coordinates of one of 30 most populous cities per country (30 per state in the United States), and due to the fixed vocabulary size of CLIP's text encoder which does not have separate tokens for many of the world's cities as these rarely occur in common text corpora.

Nevertheless, the fact that StreetCLIP was evaluated in zero-shot and can still achieve SOTA performance on an out-of-distribution benchmark signals a strong potential for using StreetCLIP as a backbone for supervised image geolocalization models as well as for finetuning StreetCLIP to perform tasks in other domains of downstream applications. 

\section{Conclusion and Future Work}

In conclusion, our experiments validate the central hypothesis of our work that synthetic caption domain-specific pretraining improves CLIP's generalized zero-shot capabilities applied to the task of image geolocalization. Our StreetCLIP model – pretrained using this method – not only improves CLIP's image geolocalization performance by 1.3 to 3.4 percentage points depending on the threshold of prediction granularity used but also achieves SOTA performance on two open-domain image geolocalization benchmarks. Notably, StreetCLIP even outcompetes models in zero-shot that were finetuned in a supervised setting on more than 4 million images originating from a distribution similar to that of our benchmarks. 

Our results have broader implications: because our synthetic caption pretraining method does not restrict StreetCLIP to make predictions within a fixed set of classes, StreetCLIP can be adapted to many other tasks benefiting from geographic domain knowledge. A wide array of potential downstream applications remain unexplored, especially in the fields of climate change mitigation, rural and urban scene understanding and education. Since our work focuses on developing a method to improve generalized zero-shot learning capabilities, we expect that finetuning our publicly available version of StreetCLIP on new datasets and problem domains could yield significant performance gains over the status quo. 

Our results further suggest that domain-specific pretraining via our synthetic caption method could potentially drive substantial prediction improvements in other domains. StreetCLIP is not restricted to image geolocalization because it only relies on class labels which can be formulated in natural language and on corresponding images. We hope that more domain-specific CLIP variants will be trained for other fields, with synthetic caption domain-specific pretraining leading to performance gains far beyond the realm of image geolocalization.

\bibliography{paper}  






\newpage
\appendix

\section{Training Details}
\label{app:training}

Table \ref{table:trainargs} presents the training arguments used in training StreetCLIP on our original datasets.

\begin{table*}[htb]
\caption{Domain-Specific Pretraining Parameters for StreetCLIP}
\label{table:trainargs}
\vskip 0.15in
\begin{center}
\begin{small}
\begin{sc}
\begin{tabular}{lc}
\toprule
\textbf{Parameter Name} & \textbf{Parameter Value} \\
\midrule
GPU Type & NVIDIA A100 80GB\\
Number of GPUs & 3\\
Batch Size & 32\\
Gradient Accumulation Steps & 12\\
Optimizer & AdamW\\
Learning Rate & $1e^{-6}$\\
Weight Decay & $1e^{-4}$\\
Warmup & 0.6 Epochs\\
Training Epochs & 3\\
Adam $\beta_1$ & 0.9\\
Adam $\beta_1$ & 0.98\\
\bottomrule
\end{tabular}
\end{sc}
\end{small}
\end{center}
\vskip -0.1in
\end{table*}

\section{Reproducing Results}

The following resources can be used to reproduce the results of this work.

\subsection{Training Dataset}
\label{app:trainingdataset}

The dataset of coordinate pairs and Street View images used to train StreetCLIP can not be released because of agreements with Geoguessr and Google.

The list of 101 countries included in our training dataset is the following:

\textit{'Albania',
 'Andorra',
 'Argentina',
 'Australia',
 'Austria',
 'Bangladesh',
 'Belgium',
 'Bermuda',
 'Bhutan',
 'Bolivia',
 'Botswana',
 'Brazil',
 'Bulgaria',
 'Cambodia',
 'Canada',
 'Chile',
 'China',
 'Colombia',
 'Croatia',
 'Czech Republic',
 'Denmark',
 'Dominican Republic',
 'Ecuador',
 'Estonia',
 'Finland',
 'France',
 'Germany',
 'Ghana',
 'Greece',
 'Greenland',
 'Guam',
 'Guatemala',
 'Hungary',
 'Iceland',
 'India',
 'Indonesia',
 'Ireland',
 'Israel',
 'Italy',
 'Japan',
 'Jordan',
 'Kenya',
 'Kyrgyzstan',
 'Laos',
 'Latvia',
 'Lesotho',
 'Lithuania',
 'Luxembourg',
 'Macedonia',
 'Madagascar',
 'Malaysia',
 'Malta',
 'Mexico',
 'Monaco',
 'Mongolia',
 'Montenegro',
 'Netherlands',
 'New Zealand',
 'Nigeria',
 'Norway',
 'Pakistan',
 'Palestine',
 'Peru',
 'Philippines',
 'Poland',
 'Portugal',
 'Puerto Rico',
 'Romania',
 'Russia',
 'Rwanda',
 'Senegal',
 'Serbia',
 'Singapore',
 'Slovakia',
 'Slovenia',
 'South Africa',
 'South Korea',
 'Spain',
 'Sri Lanka',
 'Swaziland',
 'Sweden',
 'Switzerland',
 'Taiwan',
 'Thailand',
 'Tunisia',
 'Turkey',
 'Uganda',
 'Ukraine',
 'United Arab Emirates',
 'United Kingdom',
 'United States',
 'Uruguay'}.

\subsection{StreetCLIP}

We release StreetCLIP's model weights for non-commercial purposes under the Creative Commons Attribution-NonCommercial 4.0 International license on \href{https://huggingface.co/geolocal/StreetCLIP}{HuggingFace}\footnote{Available online: \href{https://huggingface.co/geolocal/StreetCLIP}{https://huggingface.co/geolocal/StreetCLIP}.}.

\subsection{Data Required for Hierarchical Linear Probing}

\subsubsection{Comprehensive List of Countries}
\label{app:listcountries}

The following list of 193 countries was used to generate our hierarchical linear probes:

\textit{'Afghanistan',
 'Albania',
 'Algeria',
 'Andorra',
 'Angola',
 'Antigua and Barbuda',
 'Argentina',
 'Armenia',
 'Aruba',
 'Australia',
 'Austria',
 'Azerbaijan',
 'Bahamas',
 'Bahrain',
 'Bangladesh',
 'Barbados',
 'Belarus',
 'Belgium',
 'Belize',
 'Benin',
 'Bhutan',
 'Bolivia',
 'Bosnia and Herzegovina',
 'Botswana',
 'Brazil',
 'Brunei',
 'Bulgaria',
 'Burkina Faso',
 'Burundi',
 'Cabo Verde',
 'Cambodia',
 'Cameroon',
 'Canada',
 'Central African Republic',
 'Chad',
 'Chile',
 'China',
 'Colombia',
 'Comoros',
 'Costa Rica',
 'Croatia',
 'Cuba',
 'Cyprus',
 'Czech Republic',
 "Côte d'Ivoire",
 'Democratic Republic of the Congo',
 'Denmark',
 'Djibouti',
 'Dominica',
 'Dominican Republic',
 'East Timor',
 'Ecuador',
 'Egypt',
 'El Salvador',
 'Equatorial Guinea',
 'Eritrea',
 'Estonia',
 'Ethiopia',
 'Fiji',
 'Finland',
 'France',
 'Gabon',
 'Gambia',
 'Georgia',
 'Germany',
 'Ghana',
 'Greece',
 'Greenland',
 'Grenada',
 'Guatemala',
 'Guinea',
 'Guinea-Bissau',
 'Guyana',
 'Haiti',
 'Honduras',
 'Hungary',
 'Iceland',
 'India',
 'Indonesia',
 'Iran',
 'Iraq',
 'Ireland',
 'Israel',
 'Italy',
 'Jamaica',
 'Japan',
 'Jordan',
 'Kazakhstan',
 'Kenya',
 'Kuwait',
 'Kyrgyzstan',
 'Laos',
 'Latvia',
 'Lebanon',
 'Lesotho',
 'Liberia',
 'Libya',
 'Liechtenstein',
 'Lithuania',
 'Luxembourg',
 'Madagascar',
 'Malawi',
 'Malaysia',
 'Mali',
 'Malta',
 'Marshall Islands',
 'Mauritania',
 'Mauritius',
 'Mexico',
 'Micronesia',
 'Moldova',
 'Mongolia',
 'Montenegro',
 'Morocco',
 'Mozambique',
 'Myanmar',
 'Namibia',
 'Nauru',
 'Nepal',
 'Netherlands',
 'New Zealand',
 'Nicaragua',
 'Niger',
 'Nigeria',
 'North Korea',
 'North Macedonia',
 'Norway',
 'Oman',
 'Pakistan',
 'Palau',
 'Palestine',
 'Panama',
 'Papua New Guinea',
 'Paraguay',
 'Peru',
 'Philippines',
 'Poland',
 'Portugal',
 'Qatar',
 'Republic of the Congo',
 'Romania',
 'Russia',
 'Rwanda',
 'Saint Kitts and Nevis',
 'Saint Lucia',
 'Saint Vincent and the Grenadines',
 'Samoa',
 'San Marino',
 'Saudi Arabia',
 'Senegal',
 'Serbia',
 'Seychelles',
 'Sierra Leone',
 'Singapore',
 'Slovakia',
 'Slovenia',
 'Solomon Islands',
 'Somalia',
 'South Africa',
 'South Korea',
 'South Sudan',
 'Spain',
 'Sri Lanka',
 'Sudan',
 'Suriname',
 'Swaziland',
 'Sweden',
 'Switzerland',
 'Syria',
 'Taiwan',
 'Tajikistan',
 'Tanzania',
 'Thailand',
 'Togo',
 'Tonga',
 'Trinidad and Tobago',
 'Tunisia',
 'Turkey',
 'Turkmenistan',
 'Tuvalu',
 'Uganda',
 'Ukraine',
 'United Arab Emirates',
 'United Kingdom',
 'United States',
 'Uruguay',
 'Uzbekistan',
 'Vanuatu',
 'Venezuela',
 'Vietnam',
 'Yemen',
 'Zambia',
 'Zimbabwe'}.

 \subsubsection{Dataset of Cities}
 \label{app:listcities}

 We derived our dataset of city locations for hierarchical linear probing from the basic version of the World Cities Database available on \href{https://simplemaps.com/data/world-cities}{SimpleMaps}\footnote{Available online: \href{https://simplemaps.com/data/world-cities}{https://simplemaps.com/data/world-cities}.}. For linear probing, only the 30 cities with the highest population per country were used.

\subsection{Evaluation Datasets}

The following are the evaluation datasets used in our work:

\begin{itemize}
    \item IM2GPS\footnote{Available online: \href{http://graphics.cs.cmu.edu/projects/im2gps/}{http://graphics.cs.cmu.edu/projects/im2gps}.}, introduced in \citet{hays_2008}.
    \item IM2GPS3K\footnote{Available online: \href{https://github.com/lugiavn/revisiting-im2gps}{https://github.com/lugiavn/revisiting-im2gps}.}, introduced in \citet{vo_et_al}.
\end{itemize}

\end{document}